\pdfoutput=1

\documentclass[11pt]{article}

\usepackage{acl}

\usepackage{times}
\usepackage{latexsym}

\usepackage[T1]{fontenc}

\usepackage[utf8]{inputenc}

\usepackage{microtype}

\usepackage{subfigure} 
\usepackage{amsmath}
\usepackage{stfloats}
\usepackage{multirow}
\usepackage{enumitem}
\usepackage{booktabs}
\usepackage{bbding}
\usepackage{amsfonts,amssymb}
\usepackage{diagbox}
\usepackage{newfloat}
\usepackage{listings}
\usepackage{times}  
\usepackage{helvet} 
\usepackage{courier}  
\usepackage{graphicx}
\usepackage{natbib}  
\usepackage{caption}
\usepackage{makecell}

\usepackage{xcolor}

\title{ERGO: Event Relational Graph Transformer for Document-level \\ Event Causality Identification}

\author{
 Meiqi~Chen$^{1}$, Yixin~Cao$^{2}$, Kunquan~Deng$^{3}$, \\ 
 \textbf{Mukai Li$^{4}$, Kun~Wang$^{4}$, Jing~Shao$^{4}$, Yan~Zhang$^{1}$}\\ 
 $^1$ Peking University
 $^2$ Singapore Management University \\
 $^3$ Beihang University 
 $^4$ SenseTime Research \\
\texttt{meiqichen@stu.pku.edu.cn}\\
}

\begin{document}
\maketitle
\begin{abstract}
Document-level Event Causality Identification (DECI) aims to identify causal relations between event pairs in a document. It poses a great challenge of across-sentence reasoning without clear causal indicators. In this paper, we propose a novel \textbf{E}vent \textbf{R}elational  \textbf{G}raph  Transf\textbf{O}rmer (ERGO) framework for DECI, which improves existing state-of-the-art (SOTA) methods upon two aspects. First, we formulate DECI as a node classification problem by constructing an event relational graph, without the needs of prior knowledge or tools. Second, ERGO seamlessly integrates event-pair relation classification and global inference, which leverages a Relational Graph Transformer (RGT) to capture the potential causal chain. Besides, we introduce edge-building strategies and adaptive focal loss to deal with the massive false positives caused by common spurious correlation. Extensive experiments on two benchmark datasets show that ERGO significantly outperforms previous SOTA methods (13.1\% F1 gains on average). We have conducted extensive quantitative analysis and case studies to provide insights for future research directions (Section~\ref{sec:rc}).

\end{abstract}

\section{Introduction}
\label{sec:intro}
\begin{figure}
\centering  
\includegraphics[width=0.45\textwidth]{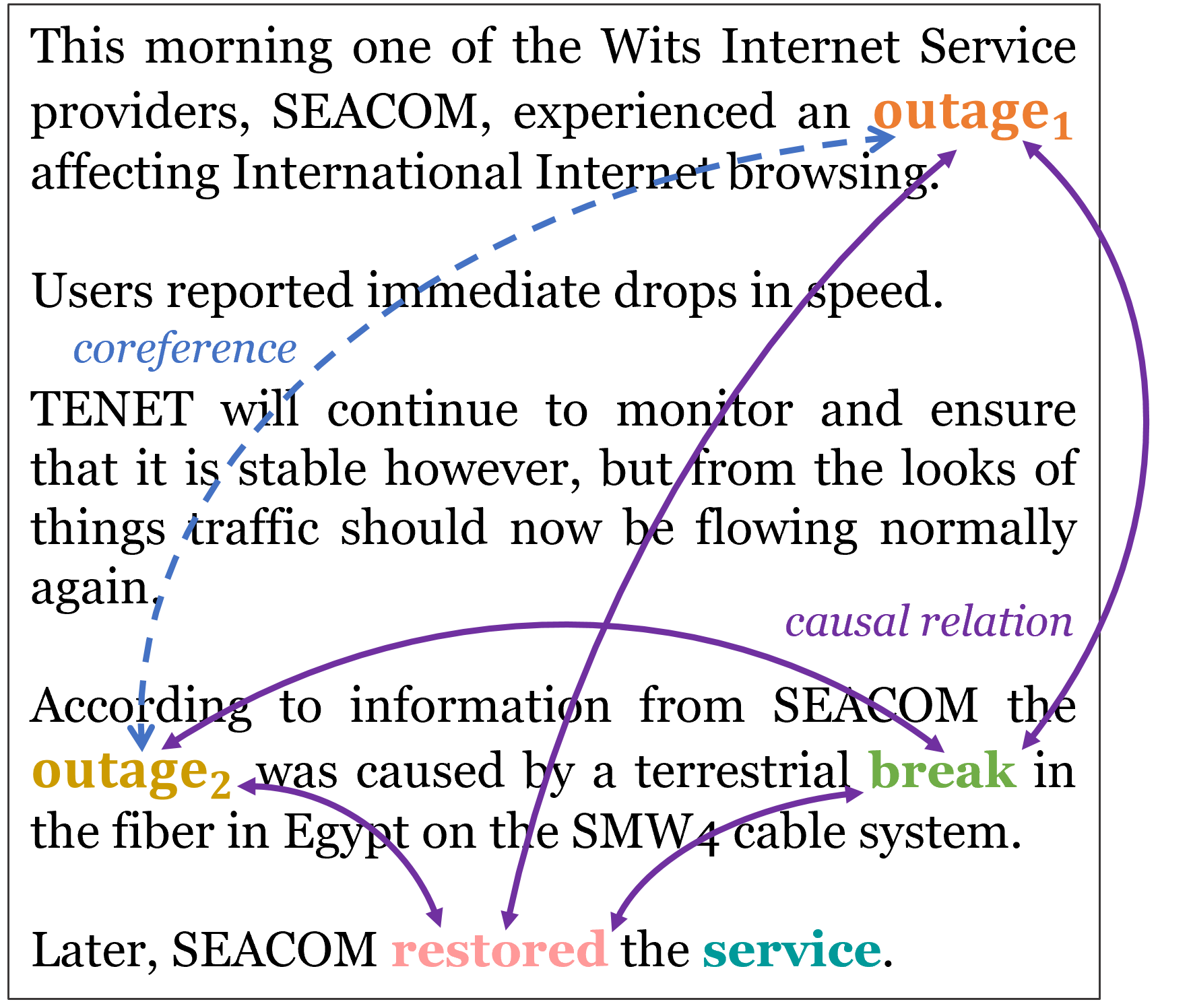}
\caption{Example of DECI. Solid purple lines denote target causal relations. The coreference relation is helpful for reasoning, denoted by the dashed blue line.}
\label{fig:example}
\vspace{-3mm}
\end{figure}

Event Causality Identification (ECI) is the task of identifying if the occurrence of one event causes another in text. As shown in Figure~\ref{fig:example}, given the text ``\emph{... the \underline{$\text{outage}_{2}$} was caused by a terrestrial \underline{break} in the fiber in Egypt ...}'', an ECI model should predict if there is a causal relation between two events triggered by ``\emph{$\text{outage}_{2}$}'' and ``\emph{break}''. Causality can reveal reliable structures of texts, which is beneficial to widespread applications, such as machine reading comprehension \citep{berant2014modeling}, question answering \citep{oh2016semi}, and future event forecasting \citep{hashimoto2019weakly}.

Existing methods mostly focus on sentence-level ECI (SECI)~\cite{liu2020knowledge, zuo2021improving, cao2021knowledge}, while in practice, a large number of causal relations are expressed with multiple sentences. This poses a new challenge of DECI -- how to conduct across-sentence reasoning without clear causal indicators~\citep{cao2021knowledge}? An event graph is typically constructed to assist the global inference, where edges are carefully designed via heuristic rules or NLP tools, such as adjacent sentences and dependency parser~\cite{gao2019modeling, phu2021graph}. However, these rules or tools are not always reliable and may introduce noise for global reasoning.

Another challenge for DECI is the imbalance of positive and negative examples -- most of the event pairs have no causal relations. In contrast,
the spurious correlation between events is more common, which may result in false-positive predictions. For example, ``treatment'' and ``death'' frequently co-occur in the same document, which may easily lead to incorrect identification of some relations between them. But, in fact, this is not causality since it is not ``treatment'' that causes ``death''. This type of negative samples are difficult to identify and particularly confuse the data-driven neural models. However, to the best of our knowledge, no work has attempted to explicitly tackle this imbalanced classification issue for ECI task.

In this paper, we propose a novel \textbf{E}vent \textbf{R}elational  \textbf{G}raph  Transf\textbf{O}rmer (ERGO) framework for DECI, which can capture potential causal chains for global reasoning. Based on "preserving transitivity of causation"~\cite{paul2013causation}, we convert ECI into a node classification problem by constructing an event relational graph, where each node denotes a pair of events, and each edge thus denotes possible transitivity among event pairs. As shown in Figure 1, if we have event ``\emph{$\text{break}$}'' causes ``\emph{$\text{outage}_{2}$}'', and ``\emph{$\text{outage}_{2}$}'' causes ``\emph{restore}''. We then can conclude ``\emph{$\text{break}$}'' causes ``\emph{restore}''. Thus, the learning on our relational graph is to capture some logic (e.g., Premise: Positive $\wedge$ Positive $\rightarrow$ Conclusion: Positive). For instance,

\noindent  \texttt{Cause(\emph{$\text{break}$}, \emph{$\text{outage}_{2}$}) $\wedge$ Cause(\emph{$\text{outage}_{2}$}, \emph{restore}) $\rightarrow$ Cause(\emph{$\text{break}$}, \emph{restore})}

or,

\noindent \texttt{Coreference(\emph{$\text{outage}_{1}$}, \emph{$\text{outage}_{2}$}) $\wedge$ Cause(\emph{$\text{break}$}, \emph{$\text{outage}_{2}$}) $\rightarrow$ Cause(\emph{$\text{break}$}, \emph{$\text{outage}_{1}$})} 

Compared with conventional event graphs, our proposed event relational graph has the following advantages. \textbf{First}, it considers high-order interactions between event pairs for reasoning: all pairs of events form a complete graph to infer transitivity relations among event relations. The easy cases (e.g., with clear causal patterns, within sentence prediction) will serve as good premises to infer the conclusion.\textbf{Second}, it does not require sophisticated graph design. Note that we do not need to know any prior relation between events. Instead, we assume each pair of events is a candidate node for binary classification --- they have causality or not. In contrast, a conventional event graph needs to carefully design the edges between events with known relations, to infer other relations between events, like a chicken and egg problem. The performance of conventional event graphs heavily depends on the structure initialization.

Nevertheless, the transitivity may be overused over the complete event relational graph, because most of the event pairs do not have causal relations. To deal with it, on the one hand, we pose a simple while effective constraint on the graph --- there is an edge between two nodes, only if they share at least one event. Otherwise, the causal chain is disconnected. On the other hand, we introduce an adaptive focal loss to deal with the imbalanced classification issue, which penalizes those hard negative samples (more discussion can be found in Section~\ref{sec:dfp}). Besides, ERGO can efficiently identify all causal relations of the document simultaneously. Our contributions can be summarized as follows:
\begin{itemize}
\item We formulate DECI as a node classification problem by building an event relational graph.
\item We propose a novel framework ERGO that models the transitivity of causation and alleviate the imbalanced classification issue to improve reasoning ability for DECI.
\item Extensive experiments on two benchmark datasets indicate that ERGO significantly outperforms the previous SOTA methods (13.1\% F1 gains on average). We have conducted extensive quantitative analysis and case studies to provide insights for future research directions (Section~\ref{sec:cs} and \ref{sec:rc}).
\end{itemize}

\section{Related Work}
ECI has attracted more and more attention in recent years. In terms of text corpus, there are mainly two types of methods: SECI and DECI.

In the first research line, early methods usually design various features tailored for causal expressions, such as lexical and syntactic patterns \citep{riaz2013toward, riaz2014depth, riaz2014recognizing}, causality cues or markers \citep{riaz2010another, do2011minimally, hidey2016identifying}, statistical information \citep{beamer2009using, hashimoto2014toward}, and temporal patterns \citep{riaz2014depth, ning2018joint}. Then, researchers resort to a large amount of labeled data to mitigate the efforts of feature engineering and to learn diverse causal expressions \citep{hu2017inference, hashimoto2019weakly}. 
To alleviate the annotation cost, recent methods leverage Pre-trained Language Models (PLMs, e.g., BERT~\citep{devlin2018bert}) for the ECI task and have achieved SOTA performance \citep{kadowaki2019event, liu2020knowledge, zuo2020knowdis}. 
To deal with implicit causal relations, \citet{cao2021knowledge} incorporate the external knowledge from ConceptNet \citep{speer2017conceptnet} for reasoning, which achieves promising results. 
\citet{zuo2021improving} learn context-specific causal patterns from external causal statements and incorporate them into a target ECI model.
\citet{zuo2021learnda} propose a data augmentation method to further solve the data lacking problem.

Along with the success of sentence-level natural language understanding, many tasks are extended to the entire document, such as relation extraction \citep{yao2019docred}, natural language inference \citep{yin2021docnli}, and event argument extraction \citep{li2021document}. DECI not only aggravates the lack of clear causal indicators but also poses the new challenge of cross-sentence inference.
\citet{gao2019modeling} use Integer Linear Programming (ILP) to model the global causal structures; RichGCN \citep{phu2021graph} constructs document-level interaction graphs and uses Graph Convolutional Network (GCN, \citet{kipf2016semi}) to capture relevant connections. 
However, the construction of the aforementioned global structure or graph requires sophisticated feature extraction or tools, which may introduce noise and mislead the model \citep{phu2021graph}. Compared with them, we formulate DECI as an efficient node classification framework, which can capture the global interactions among events automatically, as well as alleviate the common false-positive issues.
\vspace{-2mm}
\section{Methodology}
The goal of our proposed ERGO is to capture potential causal chains for document-level reasoning. There are three main components: (1) \textbf{Document Encoder} to encode the long text and output their contextualized representations; (2) \textbf{Event Relational Graph Transformer} that builds upon a handy event relational graph. It initializes node features based on outputs of the document encoder and conducts global reasoning for final causal relation classification; and (3) \textbf{Adaptive Focal Loss} that is introduced into a classifier to mitigate the dominant-negative issue.
\vspace{-2mm}
\subsection{Document Encoder}
\label{mt:enc}
Given a document $\mathcal{D} = [x_{t}]_{t=1}^{L}$ (can be of any length $L$), the document encoder aims to output the contextualized document and event representations.
We leverage a PLM as a base encoder to obtain the contextualized embeddings.
Following conventions, we add special tokens at the start and end of $\mathcal{D}$ (e.g., ``[CLS]'' and ``[SEP]'' of BERT~\citep{devlin2018bert}), and insert additional special tokens  ``<t>'' and ``</t>'' at the start and end of all events to mark the event positions. Then, we have:
\begin{equation}
\label{eq:enc}
    H = [h_{1}, h_{2}, ..., h_{L}] = \mathrm{Encoder} ([x_{1}, x_{2}, ..., x_{L}]),
\end{equation}
where $h_{i} \in \mathbb{R}^{d} $ is the embedding of token $x_{i}$. Following~\citet{soares2019matching}, we use the embedding of token ``<t>'' for event representation. 

In this paper, we choose pre-trained BERT~\citep{devlin2018bert} and Longformer~\citep{beltagy2020longformer} as encoders for comparison. We handle documents longer than the limits of PLMs as follows.

\paragraph{BERT for Document Encoder}
To handle documents that are longer than 512 (BERT's original limit), we leverage a \emph{dynamic window} to encode the entire document. Specifically, we divide $\mathcal{D}$ into several overlapping spans according to a specific step size and input them into BERT separately. To obtain the final document or event representations, we find all the embeddings of ``[CLS]'' or ``<t>'' of different spans and average them, respectively.

\paragraph{Longformer for Document Encoder}
Longformer~\citep{beltagy2020longformer} introduces a localized sliding window based attention (the default window size is 512) with
little global attention to reduce computation and extend BERT for long documents. In our implementation, we apply the efficient local and global attention pattern of Longformer. Specifically, we use global attention on the ``<s>'' token (Longformer uses ``<s>'' and ``</s>'' as the special start and end tokens, corresponding to BERT's ``[CLS]'' and ``[SEP]''), and local attention on other tokens, which could build full sequence representations. The maximum document length allowed by Longformer is 4096, which is suitable for most documents. Therefore, we directly take the embeddings of ``<s>''  as global representations and take the embeddings of ``<t>'' as event representations.

\subsection{Event Relational Graph Transformer}
\label{mt:csi}
Given contextualized embedding of each event output by the document encoder, this module first obtains event pair embeddings, then conducts further interactions among them using an event relational graph for causal relation prediction. In this section, we first introduce how to construct the relational graph, followed by modeling the graph structure information for enhanced event pair embeddings.

\subsubsection{Event Relational Graph Construction}
\label{sec:rgi}
Given a document $\mathcal{D}$ and all the events, we construct an event relational graph $\mathcal{G} = \left \{ \mathcal{V}, \mathcal{E} \right \}$, where $\mathcal{V}$ is the set of nodes, $\mathcal{E}$ is the set of edges.
We highlight the following differences of $\mathcal{G}$ from previous event graphs.
\textbf{First}, each node in $ \mathcal{V}$ refers to a different pair of events in $\mathcal{D}$, instead of a single event. Our motivation is to learn the relation of relations between events, i.e., the logic of causal transitivity, for higher-order reasoning. Some concrete examples can be found in Section~\ref{sec:intro}. Note that we do not need the exact relation, but take it as implicit knowledge to predict. We highlight coreference to denote some helpful relations, although they are not the current learning objective. We will explore them as auxiliary tasks in the future. 

\textbf{Second}, there are no pre-requirements to obtain the edges. We can simply initialize it as a complete graph, or, use the following strategy:
there is an edge between two nodes \emph{only} if the two corresponding event pairs share at least one event.
The basic idea behind this constraint is that if two pairs of events have no common event, there must be no \emph{direct} causal effect between them. That is, they have causal interactions only if there are some mediator events, and such causality takes effects conditioned on the mediator. 
For example in Figure \ref{fig:example},
(1) the causality information of event pair (\emph{restore}, \emph{service}) has no effect on predicting the causal relation of  (\emph{$\text{outage}_{1}$}, \emph{break}).
(2) the causality of (\emph{$\text{outage}_{2}$}, \emph{restored}) has a transitive effect on predicting the causal relation of  (\emph{$\text{outage}_{1}$}, \emph{break}) only if we know that (\emph{$\text{outage}_{1}$}, \emph{$\text{outage}_{2}$}) is coreference and (\emph{restored}, causes, \emph{break}).
Such high-order connectivity can still be reached by using more network layers. In Section~\ref{sec:ablation},  we compare the two edge-building strategies and results show that such a simple and intuitive constraint brings considerable performance gains.

\subsubsection{Relational Graph Transformer (RGT)}
\label{sec:rgt}
\paragraph{Event Pair Node Embeddings}
Given contextual embeddings of events output by Equation  (\ref{eq:enc}), we first initialize each event pair node embedding. Specifically, for events ($e_{1}, e_{2}$) and the corresponding contextual embeddings ($h_{e_{1}}, h_{e_{2}}$), their event pair node embedding is initialized by:
\begin{equation}
    v_{e_{1, 2}}^{(0)} = [h_{e_{1}} \| h_{e_{2}}],
\end{equation}
where $v_{e_{1, 2}}  \in \mathbb{R}^{2d}$ represents the implicit relational information between events $e_{1}$ and $e_{2}$, $\|$ denotes concatenation.

Through our event-pair nodes, we seamlessly integrate event-pair representation learning and causal chain inference, without any prior knowledge or tools. On the one hand, the structural reasoning benefits node classification; on the other hand, better event-pair representation learning will also provide a stronger premise for inference.

In each layer $l$, RGT takes a set of node embeddings $\mathbf{v}^{(l-1)} \in \mathbb{R}^{N \times d_{\mathrm{in}}}$ as input, and outputs a new set of node embeddings: $\mathbf{v}^{(l)} \in \mathbb{R}^{N \times d_{\mathrm{out}}}$, where $N$ is the number of event pairs, $d_{\mathrm{in}}$ and $d_{\mathrm{out}}$ are the dimensions of input and output embeddings.

For an event pair node $i$, to measure the importance of neighbor $j$'s relational information to node $i$, we perform a node-shared self-attention mechanism:
\begin{equation}
    {\rm att}_{ij} =\frac{ (v_{i}\mathbf{W}_{q})(v_{j}\mathbf{W}_{k})^{T}}{\sqrt{d_{k}}},
\end{equation}
where $d_{k}$ is the hidden size, $\mathbf{W}_{q}, \mathbf{W}_{k}\in \mathbb{R}^{d_{\mathrm{in}} \times d_{k}}$ are parameter weight matrices, $\sqrt{d_{k}}$ is a scaling factor \citep{vaswani2017attention}.

To make the importance more comparable, we normalize the attention coefficients across all choices of $j$ using softmax function, $\alpha_{ij}$ can reflect the contribution of neighbor $j$ to node $i$:
\begin{equation}
\label{eq:att}
\begin{aligned}
    \alpha_{ij} &=\mathrm{softmax}_{j}({\rm att}_{ij})=\frac{\exp ({\rm att}_{ij})}{\sum_{z \in \mathcal{N}_{i}} \exp ({\rm att}_{iz})},
\end{aligned}
\end{equation}
where $\mathcal{N}_{i}$ are all the first order neighbors of node $i$. 

Once we obtained the normalized attention coefficients $\alpha_{ij}$, we compute a weighted linear combination of the embeddings to aggregate relational knowledge from the neighborhood information:
\begin{equation}
    v_{i}^{(l)}=\sum_{j \in \mathcal{N}_{i}}  \alpha_{ij} (v_{j}\mathbf{W}_{v}),
\end{equation}
where $\mathbf{W}_{v}\in \mathbb{R}^{d_{\mathrm{in}} \times d_{k}}$ is the parameter weight matrix. 
We also perform multi-head attention to jointly attend to information from different representation subspaces as in \citep{vaswani2017attention}. Finally, the output embedding of node $i$ can be represented as:
\begin{equation}
    v_{i}^{(l)}= \Big ( \mathop{\Big{\|}}\limits_{c=1}^{C}\sum_{j \in \mathcal{N}_{i}}  \alpha_{ij} (v_{j}\mathbf{W}_{v})\Big )\mathbf{W}_{o},
\end{equation}
where $\|$ denotes concatenation, $C$ is the number of heads. $\mathbf{W}_{o}\in \mathbb{R}^{C\dot d_{k} \times d_{\mathrm{out}}}$ is the parameter weight matrix. By simultaneously computing embeddings for all the event pair nodes, a node embedding matrix $\mathbf{v}^{(l)} \in \mathbb{R}^{N \times d_{\mathrm{out}}}$ is obtained.

Note that \textbf{(1)} our proposed framework is flexible to almost arbitrary Graph Neural Networks (GNNs). Here we leverage RGT for its powerful expressiveness. We also report results with GCN in Section~\ref{sec:ablation}.  \textbf{(2)} We understand there are some conditions to the ``preserving transitivity''~\cite{paul2013causation}, such as exogenous variants may break the causal path, i.e., false-positive relations between events. Thus we rely on an automatic learning scheme, except for the edge constraint (Section~\ref{sec:rgi}), to deal with such hard negatives.

\subsection{Binary Classification with Focal Loss}
\vspace{-1mm}
\label{mt:bfl}
This section introduces an adaptive focal loss to mitigate the false positive issue. Remember that we formulate DECI as a node (binary) classification task, which predicts the label of each node with a positive or negative class. How can we know the difficulties of sample prediction, so that ERGO can penalize them? Since there are no annotations. What is worse, the number of negative samples during training far exceeds that of positives (causal relations are much less than the spurious correlations). This leads to an imbalanced classification problem~\citep{lin2017focal}.

To address this problem, we leverage an adaptive loss function for training, following focal loss \citep{lin2017focal}. Specifically, we reshape the loss function to down-weight easy samples and thus focus on hard ones.
Formally, a modulating factor is added to Cross-Entropy (CE) loss, with a tunable focusing parameter $\gamma \geq 0$, which is defined as:
\begin{equation}
\mathcal{L}_{\mathrm{FL}}= -\sum_{e_{i}, e_{j} \in \mathcal{D}}  (1-p_{e_{i, j}})^{\gamma}  \log (p_{e_{i, j}}).
\end{equation}
where $p_{e_{i, j}}$ is the predicted probability of whether there is a causal relation between events $e_{i}$ and $e_{j}$. $p_{e_{i, j}}$ is defined as follows:
\begin{equation}
p_{e_{i, j}} = \mathrm{softmax} \left( \left[v_{e_{i, j}} || h_{[CLS]}\right] \mathbf{W}_{p} \right),
\end{equation}
where $\mathbf{W}_{p}$ is the parameter weight matrix, $\|$ denotes concatenation. Here we concatenate embeddings of ``[CLS]'' (of BERT) or ``<s>'' (of Longformer) to each node $v_{e_{i, j}}$ in order to incorporate the global context representation for classification. 

This scaling factor, $(1-p_{e_{i, j}})^{\gamma}$, not only allows us to efficiently train on all event pairs without sampling, but also focuses the model on harder samples, thus reducing false predictions. For example, when a sample is misclassified and $p_{e_{i, j}}$ is small, the modulating factor is near 1, and the loss is unaffected. As $p_{e_{i, j}} \rightarrow$ 1, the factor goes to 0 and the loss for well-classified examples is down-weighted. Therefore, the focusing parameter $\gamma$ smoothly adjusts the rate at which easy examples are down-weighted. When $\gamma = 0$, $\mathcal{L}_{\mathrm{FL}}$ is equivalent to CE loss, and with the increase of $\gamma$, the influence of the modulating factor also increases. We will give further discussion in Section~\ref{sec:dfp}.

Besides, we use an $\alpha$-balanced variant of the focal loss, which introduces a weighting factor $\alpha$ in $[0, 1]$ for class ``positive'' and $1 - \alpha$ for class ``negative''. The value of $\alpha$ is related to the ratio of positive and negative samples.
The final adaptive focal loss $\mathcal{L}_{\mathrm{FL}_{b}}$ can be written as:
\begin{equation}
\mathcal{L}_{\mathrm{FL}_{\text{b}}}= -\sum_{e_{i}, e_{j} \in \mathcal{D}} \alpha_{e_{i, j}} (1-p_{e_{i, j}})^{\gamma}  \log (p_{e_{i, j}}).
\end{equation}

\section{Experiments}
\subsection{Datasets and Evaluation Metrics}
We evaluate our proposed method on two widely used datasets, EventStoryLine (version 0.9) \citep{caselli2017event} and Causal-TimeBank \citep{mirza2014extracting}.
\paragraph{EventStoryLine} contains 22 topics, 258 documents, 5,334 events, 7,805 intra-sentence and 62,774 inter-sentence event pairs (1,770 and 3,885 of them are annotated with causal relations respectively). 
Following \citet{gao2019modeling}, we group documents according to their topics. Documents in the last two topics are used as the development data, and documents in the remaining 20 topics are employed for a 5-fold cross-validation.
\paragraph{Causal-TimeBank} contains 184 documents, 6,813 events, and 318 of 7,608 event pairs are annotated with causal relations.
Following \citep{liu2020knowledge} and \citep{phu2021graph}, we employ a 10-fold cross-validation evaluation. Note that the number of inter-sentence event pairs in Causal-TimeBank is quite small (i.e., only 18 pairs), following \citep{phu2021graph}, we only evaluate ECI performance for intra-sentence event pairs on Causal-TimeBank.

\paragraph{Evaluation Metrics} For evaluation, we adopt Precision (P), Recall (R) and F1-score (F1) as evaluation metrics, same as previous methods~\citep{gao2019modeling,phu2021graph} to ensure comparability. 

\subsection{Implementation Details} We implement our method based on Pytorch. We use uncased BERT-base \cite{devlin2018bert} or Longformser-base \citep{beltagy2020longformer} as the document encoder.  We optimize our model with AdamW \citep{loshchilov2017decoupled} with a linear warm-up. We tune the hyper-parameters by grid search based on the development set performance and perform early stopping.

\begin{table}
    \renewcommand
    \arraystretch{1.0}
    \centering
    \small
    \setlength{\tabcolsep}{3.5pt}
        \begin{tabular}{l|ccc|ccc}
        \toprule
         \multirow{2}{*}{\bf{Model}} & \multicolumn{3}{c|}{\textbf{EventStoryLine}} & \multicolumn{3}{c}{\textbf{Causal-TimeBank}}  \\ 
         \cmidrule(lr){2-4}\cmidrule(l){5-7}
          & P & R & F1   & P & R & F1  \\ \midrule
          OP &22.5 &\textbf{98.6} &36.6 &- &- &-\\
         LR+  &37.0 &45.2 &40.7 &- &- &-        \\
         LIP  &38.8 &52.4 &44.6 &- &- &-\\ 
         \midrule
         KMMG$[\circ]$ &41.9 &62.5 &50.1 &36.6 &55.6 & 44.1 \\ 
         KnowDis$[\circ]$  &39.7 & 66.5 &49.7 &42.3 &60.5 &49.8\\
         LSIN$[\circ]$ &47.9 &58.1 &52.5 &51.5 &56.2 &52.9\\
         LearnDA$[\circ]$ &42.2 &69.8 &52.6 &41.9 &\underline{68.0} & 51.9\\
         CauSeRL$[\circ]$ &41.9 &69.0 &52.1 &43.6 &\textbf{68.1} &53.2 \\
         \midrule
         BERT$[\circ]$ &47.8 &57.2 &52.1 &47.6 &55.1 &51.1 \\
         RichGCN$[\circ]$  &49.2 &63.0 &55.2 &39.7 &56.5 &46.7 \\
         \midrule
        ERGO$[\circ]$   &\underline{49.7}  &\underline{72.6}  &\underline{59.0} &\underline{58.4} &60.5 &\underline{59.4}  \\ 
        ERGO$[\diamondsuit]$  &\textbf{57.5 } &72.0 &\textbf{63.9} &\textbf{62.1} &61.3 &\textbf{61.7}  \\ 
        \bottomrule
        \end{tabular}
       \caption{ \label{tab:eslres-intra} Model's intra-sentence performance on EventStoryLine and Causal-TimeBank, the best results are in \textbf{bold} and the second-best results are \underline{underlined}. $[\circ]$ and $[\diamondsuit]$ denote models that use pre-trained BERT-base and Longformer-base encoders, respectively. Overall, our ERGO outperforms previous SOTA models (with a significant test at the level of 0.05).}
\end{table}
\subsection{Baselines}
We compare our proposed ERGO with various state-of-the-art SECI and DECI methods.
\paragraph{SECI Baselines}
(1) \textbf{KMMG} \citep{liu2020knowledge}, which proposes a mention masking generalization method and use extenal knowledge databases.
(2) \textbf{KnowDis} \citep{zuo2020knowdis}, a knowledge enhanced distant data augmentation method to alleviate the data lacking problem.
(3) \textbf{LSIN} \citep{cao2021knowledge}, which constructs a descriptive graph to leverage external knowledge and has the current SOTA performance for intra-sentence ECI.
(4) \textbf{LearnDA} \citep{zuo2021learnda}, which uses knowledge bases to augment training data.
(5) \textbf{CauSeRL} \citep{zuo2021improving}, which learns context-specific causal patterns from external causal statements for ECI.
\paragraph{DECI Baselines}
(1) \textbf{OP} \citep{caselli2017event}, a dummy model that assigns causal relations to event pairs.
(2) \textbf{LR+} and \textbf{LIP} \citep{gao2019modeling}, feature-based methods that construct document-level structures and use various types of resources.
(3) \textbf{BERT (our implement)} a baseline method that leverages dynamic window and event marker techniques.
(4) \textbf{RichGCN} \citep{phu2021graph}, which constructs document-level interaction graph and uses GCN to capture relevant connections. RichGCN has the current SOTA performance for inter-sentence ECI.

\subsection{Overall Results}
Since some baselines are evaluated only on the EventStoryLine dataset, the baselines used for EventStoryLine and Causal-TimeBank are different. Some baselines can not handle the inter-sentence scenarios in the EventStoryLine dataset. Thus we report results of intra- and inter- sentence settings separately. 

\begin{table}
    \renewcommand
    \arraystretch{1.0}
    \centering
    \small
    \setlength{\tabcolsep}{3.5pt}
        \begin{tabular}{l|ccc|ccc}
        \toprule
         \multirow{2}{*}{\bf{Model}} & \multicolumn{3}{c|}{\textbf{Inter-sentence}} & \multicolumn{3}{c}{\textbf{Intra + Inter}}  \\ 
         \cmidrule(lr){2-4}\cmidrule(l){5-7}
          & P & R & F1   & P & R & F1  \\ \midrule
          OP  &8.4 &\textbf{99.5} &15.6 &10.5 &\textbf{99.2} &19.0 \\
         LR+  &25.2 &48.1 &33.1 &27.9 &47.2 &35.1        \\
         LIP  &35.1 &48.2 &40.6 &36.2 &49.5 &41.9\\ 
         \midrule
         BERT$[\circ]$ &36.8 &29.2 &32.6 &41.3 &38.3 &39.7\\
         RichGCN$[\circ]$  &39.2 &45.7 &42.2 &42.6 &51.3 &46.6 \\
         \midrule
        ERGO$[\circ]$   &\underline{43.2}  &\underline{48.8} &\underline{45.8} &\underline{46.3}  &50.1 &\underline{48.1}  \\ 
        ERGO$[\diamondsuit]$  &\textbf{51.6}  &43.3 &\textbf{47.1} &\textbf{48.6} &\underline{53.4} &\textbf{50.9}  \\ 
        \bottomrule
        \end{tabular}
       \caption{\label{tab:eslres-inter} Model's inter and (intra+inter)-sentence performance on EventStoryLine.}
\end{table}
\subsubsection{Intra-sentence Evaluation}
From Table \ref{tab:eslres-intra}, we can observe that:

(1) ERGO outperforms all the baselines by a large margin on both datasets. Compared with SOTA methods, ERGO-$\text{BERT}_{\text {BASE}}$ achieves 6.9\% improvements of F1-score on EventStoryLine, and  11.7\% on Causal-TimeBank. This demonstrates the effectiveness of ERGO.

(2) The feature-based method OP achieves the highest Recall on EventStoryLine, which may be due to simply assigning causal relations by mimicking textual order of presentation. This leads to many false positives and thus a low Precision.

(3) The usage of PLMs boosts the performance.
Using $\text{Longformer}_{\text {BASE}}$ as the encoder, ERGO achieves better results than ERGO-$\text{BERT}_{\text {BASE}}$, which also achieves new SOTA. The reason may be: 1) Longformer continues pre-training from Roberta \citep{liu2019roberta}, which has been found to outperform BERT on many tasks; 2) Longformer leverages an efficient local and global attention pattern, beneficial to capture longer contextual information for inference.

\subsubsection{Inter-sentence Evaluation}
From Table \ref{tab:eslres-inter}, we can observe that:

(1) ERGO greatly outperforms all the baselines under both inter- and (intra+inter)-sentence settings, especially in terms of Precision.
The promising results demonstrate that our ERGO can make better document-level inference via event relational graph, even without prior knowledge or tools. 

(2) The overall F1-score of inter-sentence setting is much lower than that of intra-sentence, which indicates the challenge of document-level ECI.

(3) The BERT baseline performs well on intra-sentence event pairs. However, it performs much worse than LIP, RichGCN, and ERGO on inter-sentence settings, which indicates that a document-level structure or graph is helpful to capture the global interactions for prediction.

\subsection{Ablation Study}
\label{sec:ablation}
\begin{table}
    \renewcommand
    \arraystretch{1.0}
    \centering
    \small
    \setlength{\tabcolsep}{8pt}
    \begin{tabular}{l|c|c|c}
    \toprule
         \textbf{Model}  & \textbf{Intra} & \textbf{Inter} &\textbf{Intra + Inter}  \\ \midrule
         ERGO$[\circ]$ &59.0 &45.8 &48.1  \\\midrule
         $\text{ERGO}_{1}[\circ]$ &56.6 &43.5 &45.6 \\
         $\text{ERGO}_{2}[\circ]$ &58.3 &43.6 &47.3  \\
         $\text{ERGO}_{3}[\circ]$ &56.2 &41.8 &44.6  \\
         \midrule[1pt]
         ERGO$[\diamondsuit]$ &\textbf{63.9} &\textbf{47.1} &\textbf{50.9}  \\\midrule
       $\text{ERGO}_{1}[\diamondsuit]$ &61.3 &44.7 &47.1 \\
      $\text{ERGO}_{2}[\diamondsuit]$ &62.6 &45.9 &49.1  \\
      $\text{ERGO}_{3}[\diamondsuit]$ &60.7 &43.1 &46.3  \\
    \bottomrule
    \end{tabular}
       \caption{\label{tab:ablation}F1 Results of Ablation study on EventStoryLine, where $\text{ERGO}_{1}$ denotes ERGO w/ a complete graph, $\text{ERGO}_{2}$ denotes ERGO w/o the focal factor, $\text{ERGO}_{3}$ denotes ERGO w/ GCN.}
\end{table}

\begin{figure}
\centering  
\includegraphics[width=0.5\textwidth]{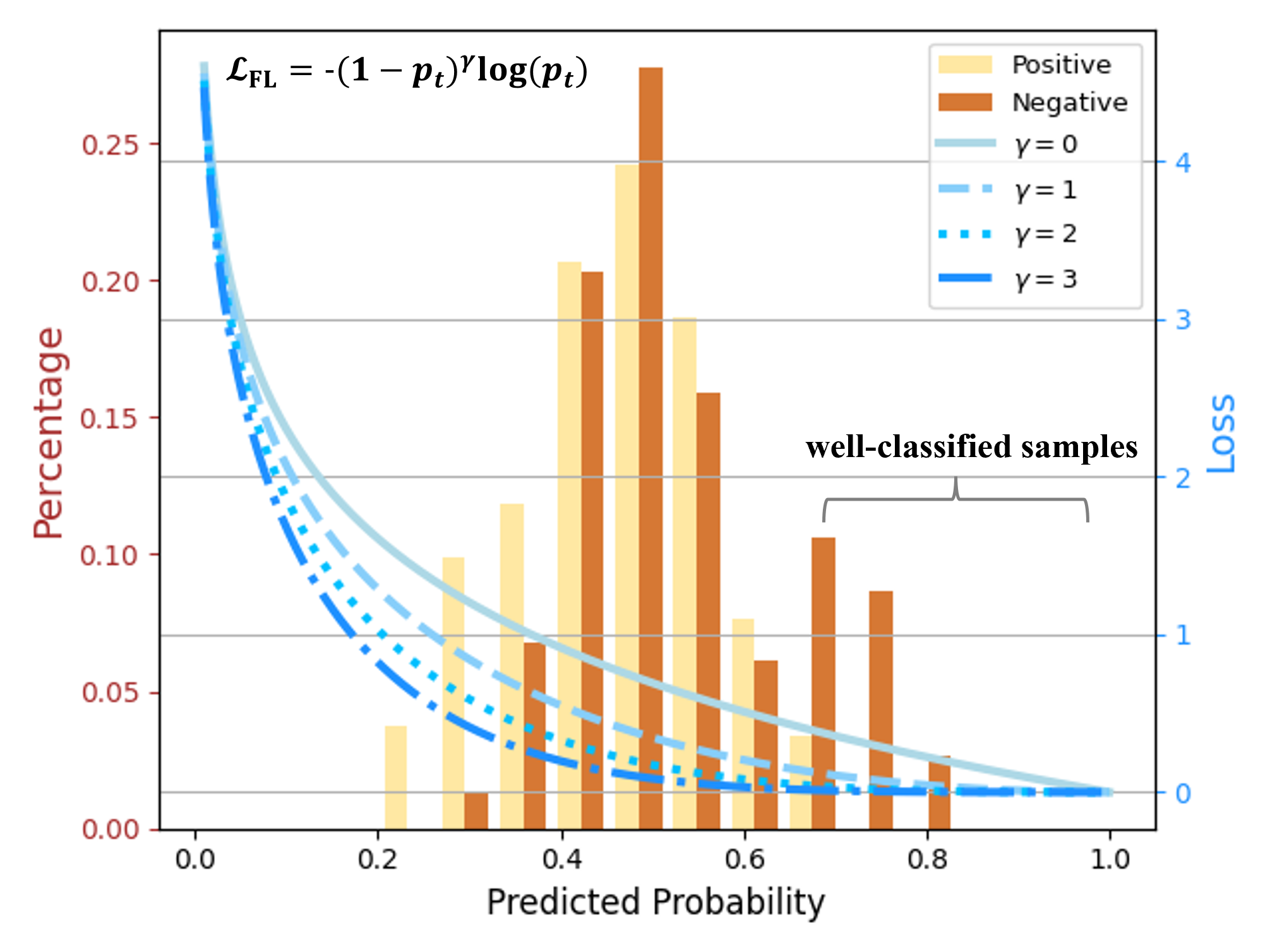}
\caption{Distribution histogram of predicted probabilities of positive and negative event pairs and the visualized loss with focal parameter $\gamma=\{0,1,2,3\}$.}
\label{fig:focal}
\end{figure}

To analyze the main components of ERGO, we have the following variants, as shown in Table \ref{tab:ablation}:

(1) \textbf{w/ a complete graph}, which connects all the nodes in the relational graph (the first edge-building strategy in Section~\ref{sec:rgi}). Comparing with the full ERGO model (both $\text{BERT}_{\text {BASE}}$ and $\text{Lonformer}_{\text {BASE}}$), ERGO (w/ a complete graph)  clearly decreases the performance, which proves the effectiveness of the handy design based on causal transitivity.

(2) \textbf{w/o focal factor}, which sets the focusing parameter $\gamma=0$ (in Section \ref{mt:bfl}), makes the focal loss degenerate into CE loss. Comparing with the full ERGO model, ERGO (w/o focal factor) also decreases the performance. This demonstrates the effectiveness of applying adaptive focal loss into ECI to deal with massive false positives.

(3) \textbf{w/ GCN}, which replaces the RGT in Section \ref{sec:rgt} with a well-known GNN model, GCN. It can be seen that \textbf{(i)} ERGO (w/ GCN) also performs better or competitive than other baselines.
This indicates that our framework is flexible to other GNNs, and the main improvement comes from our new formulation of ECI task.
\textbf{(ii)} the full ERGO model clearly outperforms ERGO (w/ GCN), which validates the effectiveness of our RGT model. 

\subsection{Dealing with False Positive}
\label{sec:dfp}

As shown in Figure \ref{fig:focal}, we show the distribution histogram of the predicted probability after the first training epoch for positive and negative samples, respectively (denoted by the bars). 
The predicted probability of x-axis reflects the difficulty of samples, and the curves denote loss --- how much penalization on the corresponding samples during learning. 
From the histogram, we can find: \textbf{(1)} the model is less confident about positives than negatives, i.e., the left-of-center distributed bars of positives. This matches our intuition that there are common spurious correlations, which brings a great challenge of the false-positive predictions (i.e., the hard negatives) to ECI.\textbf{(2)} we visualize the focal loss with $\gamma$ values $\in \{0, 1, 2, 3\}$. The top solid blue curve ($\gamma = 0$) can be seen as the standard CE loss. As $\gamma$ increases, the shape of focal loss moves to the left bottom corner. That is, the learning of ERGO pays more attention to hard negative samples. In practice, we find $\gamma = 2$ works best on both datasets, indicating that there is a balance between the focus on simple and difficult samples.

\begin{figure*}
\centering 
\includegraphics[width=1.0\textwidth]{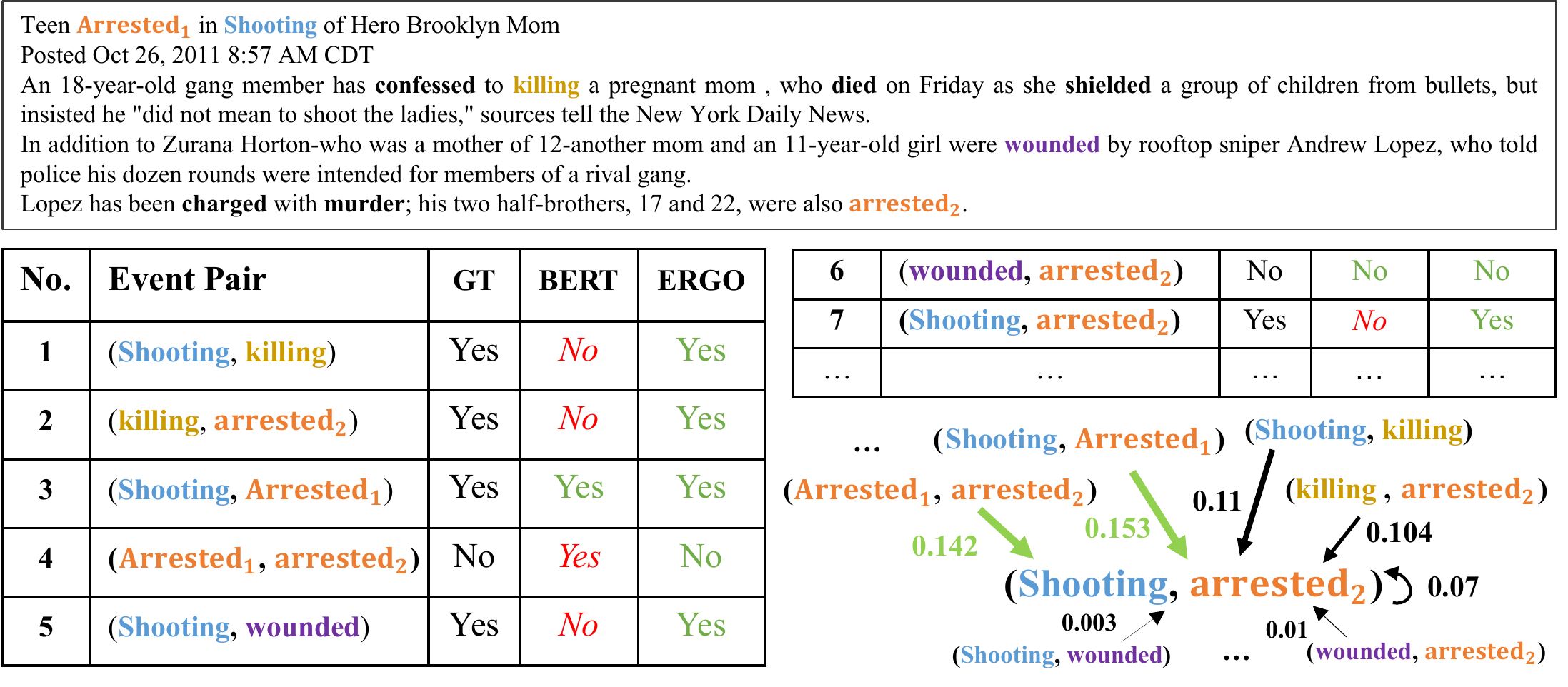} 
\caption{The case study of BERT baseline and our proposed ERGO, where ``GT'' denotes the ground truth class, and the right two columns are the output of BERT and ERGO (italic red color means wrong prediction). The thickness of arrows represents the size of attention values, and the bold green arrows show a possible reasoning path.} 
\label{fig:case} 
\end{figure*}

\begin{figure}
\centering  
\includegraphics[width=0.45\textwidth]{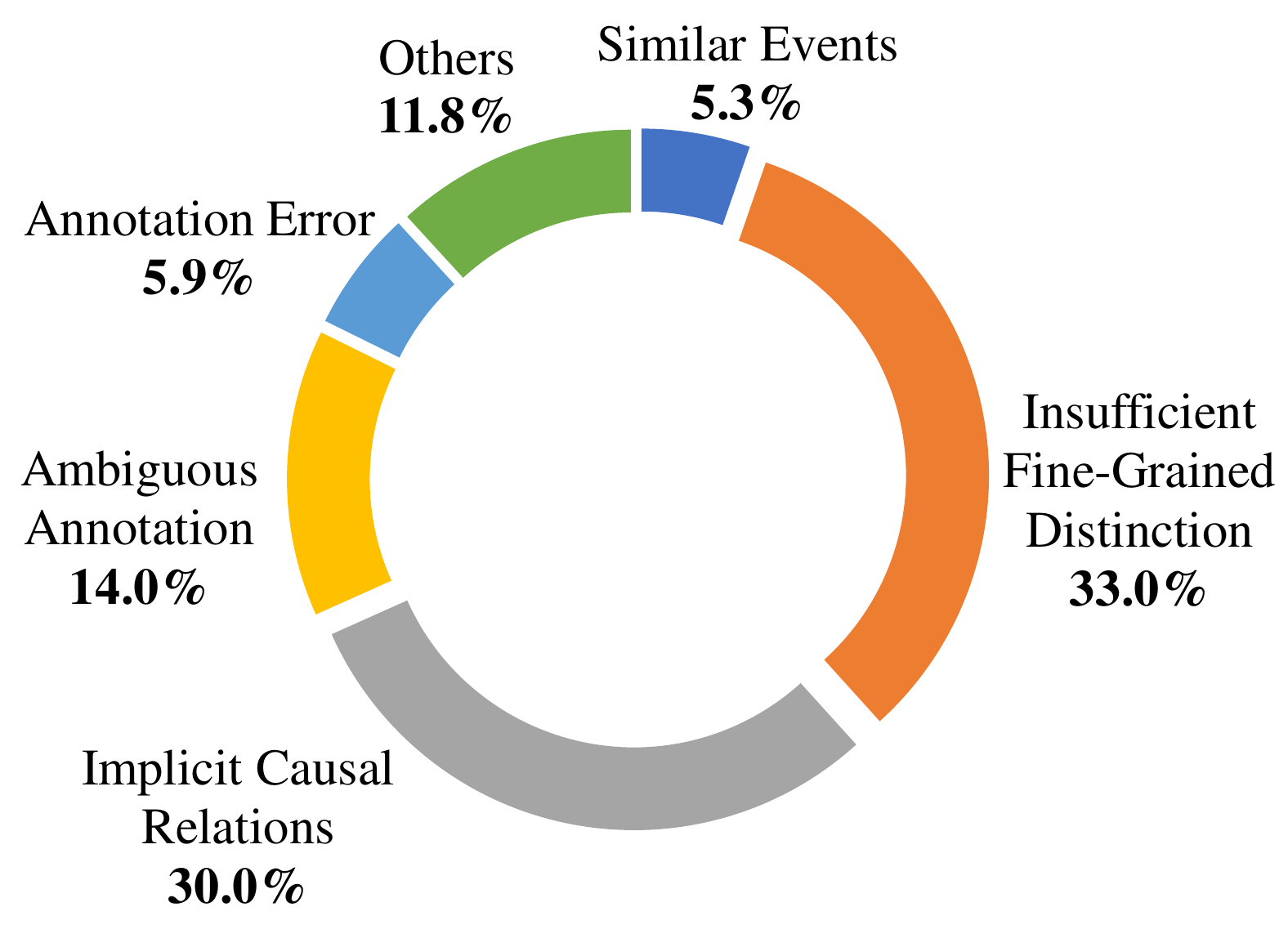}
\caption{Statistics of Error Types.}
\label{fig:error}
\end{figure}
\subsection{Case Study}
\label{sec:cs}
In this section, we conduct a case study to further illustrate an intuitive impression of our proposed ERGO. As shown in Figure \ref{fig:case}, We notice that: \textbf{(1)}
BERT is good at sentence-level ECI (e.g., No.3 event pair), but fails at more complex cross-sentence cases (e.g., No.1, 2, 4, 5, 7).
\textbf{(2)} By contrast, ERGO can make correct predictions by modeling the global interactions among event pairs. 

Figure \ref{fig:case} shows 3 causal patterns that ERGO could cover:
\textbf{(i) Transitivity} (No.1, 2, 7 event pairs): knowing both (\emph{Shooting}, \emph{killing}) and (\emph{killing}, \emph{$\text{arrested}_{2}$}) have causal relations, we could infer that (\emph{Shooting}, \emph{$\text{arrested}_{2}$}) has a causal relation;
\textbf{(ii) Implicit Coreference Assistance} (No.3, 4, 7 event pairs) : Given that (\emph{Shooting}, \emph{$\text{Arrested}_{1}$}) has a causal relation and (\emph{$\text{Arrested}_{1}$}, \emph{$\text{arrested}_{2}$}) is coreference, we could infer that (\emph{Shooting}, \emph{$\text{arrested}_{2}$}) has a causal relation, even if the causal relation of (\emph{$\text{Arrested}_{1}$}, \emph{$\text{arrested}_{2}$}) is annotated with ``No''. We attribute this to PLMs that tend to capture coreference relations, such as similar tokens. A supporting evidence is that BERT incorrectly predicts the coreferenced No.4 event pair with causal relation.
\textbf{(iii) De-confounding Spurious Correlation} (No.5, 6, 7 event pairs): Although we have (\emph{Shooting}, causes, \emph{wounded}) and (\emph{Shooting}, causes, \emph{$\text{arrested}_{2}$}), ERGO correctly recognize that there is no causal relation between (\emph{wounded}, \emph{$\text{arrested}_{2}$}). Actually, event \emph{Shooting} is the confounder that may cause spurious correlation between (\emph{wounded}, \emph{$\text{arrested}_{2}$}). This is very difficult to recognize false positives, while ERGO successfully differentiate it from causal chains by the edge constraint and adaptive focal loss (BERT also predicts correctly, may be due to it tends to predict ``No'' for cross-sentence cases, which matches the distribution of positive and negative examples).
In the bottom right, we take the event pair (\emph{Shooting}, \emph{$\text{arrested}_{2}$}) as an example and visualize the primary attention values of its neighbors computed by Equation (\ref{eq:att}). It can be seen that ERGO can learn better from informative neighbors for prediction.

\subsection{Remaining Challenges}
\label{sec:rc}
We randomly sample 20 documents of different topics from EventStoryLine, which contains 170 event pairs whose causal relations cannot be correctly predicted by our model. As shown in Figure \ref{fig:error}, we manually categorize these pairs into different types and discuss the remaining challenges:
\paragraph{Insufficient Fine-Grained Distinction and Need to Extract Temporal Information (33\%)}
For example, in the following document:

\textit{``...Dubai experienced a slight `\underline{tremor}' today, \textbf{after} a more serious \underline{earthquake} in Southern Iran, resulting in the \underline{evacuation} of Emirates Towers and a few other scrapers...''} 

The ``\emph{tremor}'' happens in ``\emph{Dubai}'' and the  ``\emph{earthquake}'' happens in  ``\emph{Southern Iran}'', they are two different events identified by the temporal indicator ``\emph{after}''. ERGO incorrectly predicts that there is a causal relation in (\emph{earthquake}, \emph{evacuation}). Future work could consider joint extraction of causal and temporal relations within the document.

\paragraph{Events with Similar Semantics (5.3\%)}
Take the following document as an example:

\textit{"...Kenneth Dorsey says the woman accused of \underline{killing} two co-workers and critically injuring a third at the Kraft plant in Northeast Philly is a good person. And so were the two women she's accused of \underline{gunning down} with a .357 Magnum, just minutes after she'd been \underline{suspended} and escorted from the building..."}

ERGO incorrectly predicts that there is a causal relation between ``\emph{killing}'' and ``\emph{gunning down}''. The reason is that ``\emph{killing}'' and ``\emph{gunning down}'' are actually coreference, which suggests a future direction in exploring related tasks.

\paragraph{Implicit Causal Relations (30\%)}
ERGO still fails at many implicit causal relations. For example, the causal relation  between ``\emph{killing}'' and ``\emph{suspended}'' in the aforementioned document. This is mainly because there are insufficient events for global reasoning and hard negatives bring noise. Clearly, commonsense reasoning will be helpful in this case, since ``\emph{suspended}'' is an unexpected change that may bring some negative emotion.
\paragraph{Ambiguous Annotation (14\%)}
This denotes those ambiguous causality within some event pairs. For example, in the following document:

 \textit{"... A Texas inmate  \underline{escaped} from a prison van near Houston after pulling a gun on two guards who were \underline{transporting} him between prisons..."}

We can think there is a causal relation between  ``\emph{escaped}'' and ``\emph{transporting}'' because if there is no ``\emph{transporting}'', the ``\emph{inmate}'' will have no chance to ``\emph{escape}''. However, we can also think that there is no causal relation between them, because it is not ``\emph{transporting}'' that directly causes ``\emph{escape}''.

Finally, our statistics shows that the other errors have to do with annotation errors (5.9\%) and more complicated issues that cannot be categorized clearly (``Others'', 11.8\%).
\section{Conclusion}
In this paper, we regard DECI as a node classification task by constructing an event relational graph. We propose a novel Event Relational Graph Transformer (ERGO) framework that could capture potential causal chains and penalize those hard negatives for DECI.
Extensive experiments show a significant improvement of ERGO for both intra- and inter-sentence ECI on two widely used benchmarks. We also conduct extensive analysis and case studies to provide insights for future research directions. In the future, we will consider introducing commonsense reasoning and auxiliary tasks to improve performance.
\bibliography{anthology,custom}

\begin{thebibliography}{36}
\expandafter\ifx\csname natexlab\endcsname\relax\def\natexlab#1{#1}\fi

\bibitem[{Baldini~Soares et~al.(2019)Baldini~Soares, FitzGerald, Ling, and
  Kwiatkowski}]{soares2019matching}
Livio Baldini~Soares, Nicholas FitzGerald, Jeffrey Ling, and Tom Kwiatkowski.
  2019.
\newblock \href {https://doi.org/10.18653/v1/P19-1279} {Matching the blanks:
  Distributional similarity for relation learning}.
\newblock In \emph{Proceedings of the 57th Annual Meeting of the Association
  for Computational Linguistics}, pages 2895--2905, Florence, Italy.
  Association for Computational Linguistics.

\bibitem[{Beamer and Girju(2009)}]{beamer2009using}
Brandon Beamer and Roxana Girju. 2009.
\newblock Using a bigram event model to predict causal potential.
\newblock In \emph{International Conference on Intelligent Text Processing and
  Computational Linguistics}, pages 430--441. Springer.

\bibitem[{Beltagy et~al.(2020)Beltagy, Peters, and
  Cohan}]{beltagy2020longformer}
Iz~Beltagy, Matthew~E Peters, and Arman Cohan. 2020.
\newblock \href {https://arxiv.org/abs/2004.05150} {Longformer: The
  long-document transformer}.
\newblock \emph{ArXiv preprint}, abs/2004.05150.

\bibitem[{Berant et~al.(2014)Berant, Srikumar, Chen, Vander~Linden, Harding,
  Huang, Clark, and Manning}]{berant2014modeling}
Jonathan Berant, Vivek Srikumar, Pei-Chun Chen, Abby Vander~Linden, Brittany
  Harding, Brad Huang, Peter Clark, and Christopher~D. Manning. 2014.
\newblock \href {https://doi.org/10.3115/v1/D14-1159} {Modeling biological
  processes for reading comprehension}.
\newblock In \emph{Proceedings of the 2014 Conference on Empirical Methods in
  Natural Language Processing ({EMNLP})}, pages 1499--1510, Doha, Qatar.
  Association for Computational Linguistics.

\bibitem[{Cao et~al.(2021)Cao, Zuo, Chen, Liu, Zhao, Chen, and
  Peng}]{cao2021knowledge}
Pengfei Cao, Xinyu Zuo, Yubo Chen, Kang Liu, Jun Zhao, Yuguang Chen, and Weihua
  Peng. 2021.
\newblock \href {https://doi.org/10.18653/v1/2021.acl-long.376}
  {Knowledge-enriched event causality identification via latent structure
  induction networks}.
\newblock In \emph{Proceedings of the 59th ACL and the 11th IJCNLP (Volume 1:
  Long Papers)}, pages 4862--4872, Online. Association for Computational
  Linguistics.

\bibitem[{Caselli and Vossen(2017)}]{caselli2017event}
Tommaso Caselli and Piek Vossen. 2017.
\newblock \href {https://doi.org/10.18653/v1/W17-2711} {The event {S}tory{L}ine
  corpus: A new benchmark for causal and temporal relation extraction}.
\newblock In \emph{Proceedings of the Events and Stories in the News Workshop},
  pages 77--86, Vancouver, Canada. Association for Computational Linguistics.

\bibitem[{Devlin et~al.(2019)Devlin, Chang, Lee, and
  Toutanova}]{devlin2018bert}
Jacob Devlin, Ming-Wei Chang, Kenton Lee, and Kristina Toutanova. 2019.
\newblock \href {https://doi.org/10.18653/v1/N19-1423} {{BERT}: Pre-training of
  deep bidirectional transformers for language understanding}.
\newblock In \emph{Proceedings of the 2019 NAACL: Human Language Technologies,
  Volume 1 (Long and Short Papers)}, pages 4171--4186, Minneapolis, Minnesota.
  Association for Computational Linguistics.

\bibitem[{Do et~al.(2011)Do, Chan, and Roth}]{do2011minimally}
Quang Do, Yee~Seng Chan, and Dan Roth. 2011.
\newblock \href {https://aclanthology.org/D11-1027} {Minimally supervised event
  causality identification}.
\newblock In \emph{Proceedings of the 2011 Conference on Empirical Methods in
  Natural Language Processing}, pages 294--303, Edinburgh, Scotland, UK.
  Association for Computational Linguistics.

\bibitem[{Gao et~al.(2019)Gao, Choubey, and Huang}]{gao2019modeling}
Lei Gao, Prafulla~Kumar Choubey, and Ruihong Huang. 2019.
\newblock \href {https://doi.org/10.18653/v1/N19-1179} {Modeling document-level
  causal structures for event causal relation identification}.
\newblock In \emph{Proceedings of the 2019 NAACL: Human Language Technologies,
  Volume 1 (Long and Short Papers)}, pages 1808--1817, Minneapolis, Minnesota.
  Association for Computational Linguistics.

\bibitem[{Hashimoto(2019)}]{hashimoto2019weakly}
Chikara Hashimoto. 2019.
\newblock \href {https://doi.org/10.18653/v1/D19-1296} {Weakly supervised
  multilingual causality extraction from {W}ikipedia}.
\newblock In \emph{Proceedings of the 2019 Conference on Empirical Methods in
  Natural Language Processing and the 9th International Joint Conference on
  Natural Language Processing (EMNLP-IJCNLP)}, pages 2988--2999, Hong Kong,
  China. Association for Computational Linguistics.

\bibitem[{Hashimoto et~al.(2014)Hashimoto, Torisawa, Kloetzer, Sano, Varga, Oh,
  and Kidawara}]{hashimoto2014toward}
Chikara Hashimoto, Kentaro Torisawa, Julien Kloetzer, Motoki Sano, Istv{\'a}n
  Varga, Jong-Hoon Oh, and Yutaka Kidawara. 2014.
\newblock \href {https://doi.org/10.3115/v1/P14-1093} {Toward future scenario
  generation: Extracting event causality exploiting semantic relation, context,
  and association features}.
\newblock In \emph{Proceedings of the 52nd ACL (Volume 1: Long Papers)}, pages
  987--997, Baltimore, Maryland. Association for Computational Linguistics.

\bibitem[{Hidey and McKeown(2016)}]{hidey2016identifying}
Christopher Hidey and Kathy McKeown. 2016.
\newblock \href {https://doi.org/10.18653/v1/P16-1135} {Identifying causal
  relations using parallel {W}ikipedia articles}.
\newblock In \emph{Proceedings of the 54th Annual Meeting of the Association
  for Computational Linguistics (Volume 1: Long Papers)}, pages 1424--1433,
  Berlin, Germany. Association for Computational Linguistics.

\bibitem[{Hu et~al.(2017)Hu, Rahimtoroghi, and Walker}]{hu2017inference}
Zhichao Hu, Elahe Rahimtoroghi, and Marilyn Walker. 2017.
\newblock \href {https://doi.org/10.18653/v1/W17-2708} {Inference of
  fine-grained event causality from blogs and films}.
\newblock In \emph{Proceedings of the Events and Stories in the News Workshop},
  pages 52--58, Vancouver, Canada. Association for Computational Linguistics.

\bibitem[{Kadowaki et~al.(2019)Kadowaki, Iida, Torisawa, Oh, and
  Kloetzer}]{kadowaki2019event}
Kazuma Kadowaki, Ryu Iida, Kentaro Torisawa, Jong-Hoon Oh, and Julien Kloetzer.
  2019.
\newblock \href {https://doi.org/10.18653/v1/D19-1590} {Event causality
  recognition exploiting multiple annotators{'} judgments and background
  knowledge}.
\newblock In \emph{Proceedings of the 2019 Conference on Empirical Methods in
  Natural Language Processing and the 9th International Joint Conference on
  Natural Language Processing (EMNLP-IJCNLP)}, pages 5816--5822, Hong Kong,
  China. Association for Computational Linguistics.

\bibitem[{Kipf and Welling(2017)}]{kipf2016semi}
Thomas~N. Kipf and Max Welling. 2017.
\newblock \href {https://openreview.net/forum?id=SJU4ayYgl} {Semi-supervised
  classification with graph convolutional networks}.
\newblock In \emph{5th International Conference on Learning Representations,
  {ICLR} 2017, Toulon, France, April 24-26, 2017, Conference Track
  Proceedings}. OpenReview.net.

\bibitem[{Li et~al.(2021)Li, Ji, and Han}]{li2021document}
Sha Li, Heng Ji, and Jiawei Han. 2021.
\newblock \href {https://doi.org/10.18653/v1/2021.naacl-main.69}
  {Document-level event argument extraction by conditional generation}.
\newblock In \emph{Proceedings of the 2021 NAACL: Human Language Technologies},
  pages 894--908, Online. Association for Computational Linguistics.

\bibitem[{Lin et~al.(2017)Lin, Goyal, Girshick, He, and
  Doll{\'{a}}r}]{lin2017focal}
Tsung{-}Yi Lin, Priya Goyal, Ross~B. Girshick, Kaiming He, and Piotr
  Doll{\'{a}}r. 2017.
\newblock \href {https://doi.org/10.1109/ICCV.2017.324} {Focal loss for dense
  object detection}.
\newblock In \emph{{IEEE} International Conference on Computer Vision, {ICCV}
  2017, Venice, Italy, October 22-29, 2017}, pages 2999--3007. {IEEE} Computer
  Society.

\bibitem[{Liu et~al.(2020)Liu, Chen, and Zhao}]{liu2020knowledge}
Jian Liu, Yubo Chen, and Jun Zhao. 2020.
\newblock \href {https://doi.org/10.24963/ijcai.2020/499} {Knowledge enhanced
  event causality identification with mention masking generalizations}.
\newblock In \emph{Proceedings of the Twenty-Ninth International Joint
  Conference on Artificial Intelligence, {IJCAI} 2020}, pages 3608--3614.
  ijcai.org.

\bibitem[{Liu et~al.(2019)Liu, Ott, Goyal, Du, Joshi, Chen, Levy, Lewis,
  Zettlemoyer, and Stoyanov}]{liu2019roberta}
Yinhan Liu, Myle Ott, Naman Goyal, Jingfei Du, Mandar Joshi, Danqi Chen, Omer
  Levy, Mike Lewis, Luke Zettlemoyer, and Veselin Stoyanov. 2019.
\newblock \href {https://arxiv.org/abs/1907.11692} {Roberta: A robustly
  optimized bert pretraining approach}.
\newblock \emph{ArXiv preprint}, abs/1907.11692.

\bibitem[{Loshchilov and Hutter(2019)}]{loshchilov2017decoupled}
Ilya Loshchilov and Frank Hutter. 2019.
\newblock \href {https://openreview.net/forum?id=Bkg6RiCqY7} {Decoupled weight
  decay regularization}.
\newblock In \emph{7th International Conference on Learning Representations,
  {ICLR} 2019, New Orleans, LA, USA, May 6-9, 2019}. OpenReview.net.

\bibitem[{Mirza(2014)}]{mirza2014extracting}
Paramita Mirza. 2014.
\newblock \href {https://doi.org/10.3115/v1/P14-3002} {Extracting temporal and
  causal relations between events}.
\newblock In \emph{Proceedings of the {ACL} 2014 Student Research Workshop},
  pages 10--17, Baltimore, Maryland, USA. Association for Computational
  Linguistics.

\bibitem[{Ning et~al.(2018)Ning, Feng, Wu, and Roth}]{ning2018joint}
Qiang Ning, Zhili Feng, Hao Wu, and Dan Roth. 2018.
\newblock \href {https://doi.org/10.18653/v1/P18-1212} {Joint reasoning for
  temporal and causal relations}.
\newblock In \emph{Proceedings of the 56th Annual Meeting of the Association
  for Computational Linguistics (Volume 1: Long Papers)}, pages 2278--2288,
  Melbourne, Australia. Association for Computational Linguistics.

\bibitem[{Oh et~al.(2016)Oh, Torisawa, Hashimoto, Iida, Tanaka, and
  Kloetzer}]{oh2016semi}
Jong{-}Hoon Oh, Kentaro Torisawa, Chikara Hashimoto, Ryu Iida, Masahiro Tanaka,
  and Julien Kloetzer. 2016.
\newblock \href
  {http://www.aaai.org/ocs/index.php/AAAI/AAAI16/paper/view/12208} {A
  semi-supervised learning approach to why-question answering}.
\newblock In \emph{Proceedings of the Thirtieth {AAAI} Conference on Artificial
  Intelligence, February 12-17, 2016, Phoenix, Arizona, {USA}}, pages
  3022--3029. {AAAI} Press.

\bibitem[{Paul et~al.(2013)Paul, Hall, and Hall}]{paul2013causation}
Laurie~Ann Paul, Ned Hall, and Edward~Jonathan Hall. 2013.
\newblock \emph{Causation: A user's guide}.
\newblock Oxford University Press.

\bibitem[{Riaz and Girju(2010)}]{riaz2010another}
Mehwish Riaz and Roxana Girju. 2010.
\newblock Another look at causality: Discovering scenario-specific contingency
  relationships with no supervision.
\newblock In \emph{2010 IEEE Fourth International Conference on Semantic
  Computing}, pages 361--368. IEEE.

\bibitem[{Riaz and Girju(2013)}]{riaz2013toward}
Mehwish Riaz and Roxana Girju. 2013.
\newblock \href {https://aclanthology.org/W13-4004} {Toward a better
  understanding of causality between verbal events: Extraction and analysis of
  the causal power of verb-verb associations}.
\newblock In \emph{Proceedings of the {SIGDIAL} 2013 Conference}, pages 21--30,
  Metz, France. Association for Computational Linguistics.

\bibitem[{Riaz and Girju(2014{\natexlab{a}})}]{riaz2014depth}
Mehwish Riaz and Roxana Girju. 2014{\natexlab{a}}.
\newblock \href {https://doi.org/10.3115/v1/W14-4322} {In-depth exploitation of
  noun and verb semantics to identify causation in verb-noun pairs}.
\newblock In \emph{Proceedings of the 15th Annual Meeting of the Special
  Interest Group on Discourse and Dialogue ({SIGDIAL})}, pages 161--170,
  Philadelphia, PA, U.S.A. Association for Computational Linguistics.

\bibitem[{Riaz and Girju(2014{\natexlab{b}})}]{riaz2014recognizing}
Mehwish Riaz and Roxana Girju. 2014{\natexlab{b}}.
\newblock \href {https://doi.org/10.3115/v1/W14-0707} {Recognizing causality in
  verb-noun pairs via noun and verb semantics}.
\newblock In \emph{Proceedings of the {EACL} 2014 Workshop on Computational
  Approaches to Causality in Language ({CA}to{CL})}, pages 48--57, Gothenburg,
  Sweden. Association for Computational Linguistics.

\bibitem[{Speer et~al.(2017)Speer, Chin, and Havasi}]{speer2017conceptnet}
Robyn Speer, Joshua Chin, and Catherine Havasi. 2017.
\newblock \href {http://aaai.org/ocs/index.php/AAAI/AAAI17/paper/view/14972}
  {Conceptnet 5.5: An open multilingual graph of general knowledge}.
\newblock In \emph{Proceedings of the Thirty-First {AAAI} Conference on
  Artificial Intelligence, February 4-9, 2017, San Francisco, California,
  {USA}}, pages 4444--4451. {AAAI} Press.

\bibitem[{Tran~Phu and Nguyen(2021)}]{phu2021graph}
Minh Tran~Phu and Thien~Huu Nguyen. 2021.
\newblock \href {https://doi.org/10.18653/v1/2021.naacl-main.273} {Graph
  convolutional networks for event causality identification with rich
  document-level structures}.
\newblock In \emph{Proceedings of the 2021 Conference of the North American
  Chapter of the Association for Computational Linguistics: Human Language
  Technologies}, pages 3480--3490, Online. Association for Computational
  Linguistics.

\bibitem[{Vaswani et~al.(2017)Vaswani, Shazeer, Parmar, Uszkoreit, Jones,
  Gomez, Kaiser, and Polosukhin}]{vaswani2017attention}
Ashish Vaswani, Noam Shazeer, Niki Parmar, Jakob Uszkoreit, Llion Jones,
  Aidan~N. Gomez, Lukasz Kaiser, and Illia Polosukhin. 2017.
\newblock \href
  {https://proceedings.neurips.cc/paper/2017/hash/3f5ee243547dee91fbd053c1c4a845aa-Abstract.html}
  {Attention is all you need}.
\newblock In \emph{Advances in Neural Information Processing Systems 30: Annual
  Conference on Neural Information Processing Systems 2017, December 4-9, 2017,
  Long Beach, CA, {USA}}, pages 5998--6008.

\bibitem[{Yao et~al.(2019)Yao, Ye, Li, Han, Lin, Liu, Liu, Huang, Zhou, and
  Sun}]{yao2019docred}
Yuan Yao, Deming Ye, Peng Li, Xu~Han, Yankai Lin, Zhenghao Liu, Zhiyuan Liu,
  Lixin Huang, Jie Zhou, and Maosong Sun. 2019.
\newblock \href {https://doi.org/10.18653/v1/P19-1074} {{D}oc{RED}: A
  large-scale document-level relation extraction dataset}.
\newblock In \emph{Proceedings of the 57th ACL}, pages 764--777, Florence,
  Italy. Association for Computational Linguistics.

\bibitem[{Yin et~al.(2021)Yin, Radev, and Xiong}]{yin2021docnli}
Wenpeng Yin, Dragomir Radev, and Caiming Xiong. 2021.
\newblock \href {https://doi.org/10.18653/v1/2021.findings-acl.435}
  {{D}oc{NLI}: A large-scale dataset for document-level natural language
  inference}.
\newblock In \emph{Findings of the Association for Computational Linguistics:
  ACL-IJCNLP 2021}, pages 4913--4922, Online. Association for Computational
  Linguistics.

\bibitem[{Zuo et~al.(2021{\natexlab{a}})Zuo, Cao, Chen, Liu, Zhao, Peng, and
  Chen}]{zuo2021improving}
Xinyu Zuo, Pengfei Cao, Yubo Chen, Kang Liu, Jun Zhao, Weihua Peng, and Yuguang
  Chen. 2021{\natexlab{a}}.
\newblock \href {https://doi.org/10.18653/v1/2021.findings-acl.190} {Improving
  event causality identification via self-supervised representation learning on
  external causal statement}.
\newblock In \emph{Findings of the Association for Computational Linguistics:
  ACL-IJCNLP 2021}, pages 2162--2172, Online. Association for Computational
  Linguistics.

\bibitem[{Zuo et~al.(2021{\natexlab{b}})Zuo, Cao, Chen, Liu, Zhao, Peng, and
  Chen}]{zuo2021learnda}
Xinyu Zuo, Pengfei Cao, Yubo Chen, Kang Liu, Jun Zhao, Weihua Peng, and Yuguang
  Chen. 2021{\natexlab{b}}.
\newblock \href {https://doi.org/10.18653/v1/2021.acl-long.276} {{L}earn{DA}:
  Learnable knowledge-guided data augmentation for event causality
  identification}.
\newblock In \emph{Proceedings of the 59th ACL and the 11th IJCNLP (Volume 1:
  Long Papers)}, pages 3558--3571, Online. Association for Computational
  Linguistics.

\bibitem[{Zuo et~al.(2020)Zuo, Chen, Liu, and Zhao}]{zuo2020knowdis}
Xinyu Zuo, Yubo Chen, Kang Liu, and Jun Zhao. 2020.
\newblock \href {https://doi.org/10.18653/v1/2020.coling-main.135}
  {{K}now{D}is: Knowledge enhanced data augmentation for event causality
  detection via distant supervision}.
\newblock In \emph{Proceedings of the 28th International Conference on
  Computational Linguistics}, pages 1544--1550, Barcelona, Spain (Online).
  International Committee on Computational Linguistics.

\end{thebibliography}
\bibliographystyle{acl_natbib}
\clearpage
\end{document}